\documentclass[10pt]{llncs}
\usepackage{rotating}
\usepackage{pdflscape}
\usepackage{amsmath}
\usepackage{amsfonts}
\usepackage{amssymb}
\usepackage{amstext}
\usepackage{latexsym,stmaryrd}
\usepackage{algorithm, algorithmic}
\usepackage{multirow}
\usepackage{inferance}
\usepackage{deduc}

\def\infty{\text{BUG}}

\renewcommand{\leq}{\mathrel{\leqslant}}
\renewcommand{\geq}{\mathrel{\geqslant}}

\newcommand{\lconj}{\ensuremath{\mathbin{\wedge}}}
\newcommand{\ldisj}{\ensuremath{\mathbin{\vee}}}

\renewcommand\ruleformat[1]{\ #1}

\begin{document}
\title{Mining to Compact CNF Propositional Formulae}
\author{Said Jabbour and Lakhdar Sais and Yakoub Salhi}
\institute{CRIL - CNRS, University of Artois, France}
\maketitle

\begin{abstract}
 In this paper, we propose a first application of 
data mining techniques to propositional satisfiability. 
Our proposed Mining4SAT approach aims to discover and to exploit hidden structural knowledge for reducing the size of propositional formulae in conjunctive normal form (CNF). 
Mining4SAT  combines  both frequent itemset mining techniques and Tseitin's
encoding for a compact representation of CNF formulae. The experiments of our Mining4SAT approach  show
 interesting reductions of the sizes of many application instances taken from the last SAT competitions.

\end{abstract}

\section{Introduction}
\label{sect:introduction}

Propositional satisfiability (SAT) became a core technology in many application domains, such as formal verification, 
planning and various new applications  derived by the recent impressive progress in practical SAT solving.  
Propositional formulae in conjunctive normal form (CNF) is the standard input format for propositional satisfiability. 
Such convenient CNF form is derived from a general boolean formula  using the well-known Tseitin encoding \cite{Tseitin68}. 
Two important flaws were identified and largely discussed in the literature. First, it is often argued that by encoding 
arbitrary propositional formulae in CNF, structural properties of the original problem are not reflected in the CNF formula. Secondly, 
even if such translation is linear in the size of the original formula, a huge CNF formula might result when encoding 
real-world problems.  Some instances exceed the capacity of the available memory, and even if the instance can be stored, 
the time needed for reading the input instance might be higher than its solving time. 

To address this problem,  developing a more compact representation is clearly an interesting research issue. 
By compact encoding of formulae, we have in mind a representation model which
through its use of structural properties results in the most compact possible formula.

Two promising models were proposed these last years. The first, proposed by H. Dixon et al \cite{DixonGHLP04}, uses group theory to
represent several classical clauses by a single clause called an "augmented clause". The second model was proposed 
by M. L. Ginsberg et al  \cite{Ginsberg00}, called QPROP ("quantified propositional logic"),
which may be seen as a propositional formula extended by the introduction of
quantifications over finite domains, i.e. first order logic limited to finite types and without 
functional symbols. The problem rises in finding efficient solving techniques  of formulae encoded using such models. 

More recently, an original approach for compacting sets of binary clauses was proposed by J. Rintanen in \cite{Rintanen06}. 
Binary clauses  are ubiquitous in 
propositional formulae that represent real-world problems ranging from model-checking 
problems in computer-aided verification to AI planning problems. In \cite{Rintanen06}, using auxiliary variables, it is shown 
how constraint graphs that contain big cliques or bi-cliques of binary clauses can be represented more compactly than the quadratic and explicit representation. 
The main limitation of this approach lies in its restriction to particular sets of binary clauses whose constraints graph represents cliques or bi-cliques. 
Such particular regularities can caused by the presence of an at-most-one constraint  over a subset of variables, forbidding more than one 
of them to be true at a time. 

In data mining community, several models and techniques for discovering interesting patterns 
in large databases has been proposed in the last few years.  The problem of mining frequent itemsets is well-known 
and essential in data mining, knowledge discovery and data analysis. Since the first article of Agrawal \cite{agrawal93} 
on association rules and itemset mining, the huge number of works, challenges, datasets and projects show the actual interest 
in this problem (see \cite{tiwari2010survey} for a recent survey). 

Our goal in this work is to address the problem of finding compact representation of  CNF formulae.
Our proposed Mining4SAT approach aims 
to {\it discover hidden structures}  from arbitrary CNF formulae and to exploit them to reduce the overall size of the CNF formula 
while preserving satisfiability. Mining4SAT makes an original use for SAT of an exciting novel application domain, namely, the data mining task 
of finding frequent itemset from 0-1 transaction databases \cite{agrawal93}.  

Recently, a first constraint programming (CP) based data mining framework was proposed  by Luc De Raedt  et al. in \cite{Raedt08} for itemset mining. 
This new framework offers a declarative and flexible representation model. It allows data mining problems to benefit from several generic and efficient CP 
solving techniques  \cite{GunsNR11}.  This first study leads to the first CP approach for itemset mining displaying nice declarative opportunities while opening 
 interesting perspectives to cross fertilization between data-mining, constraint programming and propositional satisfiability. 
 
 In this paper, we are particularly  interested in the other side of this innovative connection between these two research domains, namely how data-mining 
 can be helpful for SAT.   We present the first data-mining approach for Boolean Satisfiability. We show that itemset mining techniques are very suitable  
 for discovering interesting patterns from CNF formulae.  Such patterns are then used to rewrite the CNF formula more compactly. We also show how sets of binary clauses can be also compacted by our approach. Wa also prove that our approach can automatically achieve similar reductions as in \cite{Rintanen06}, on bi-cliques and cliques of binary clauses.   It is also important to note, that our proposed mining4SAT approach is incremental. Indeed, our method can be applied incrementally or in parallel on the subsets of any partition of the original CNF formula. This will be particularly helpful for huge CNF formula that can not be entirely stored in memory.  


\section{Frequent Itemset Mining Problem}
\label{sec:im}
\subsection{Preliminary Notations and Definitions}

Let $\cal I$ be a set of {\it items}. A set $I\subseteq {\cal I}$ is called an {\it itemset}.
A {\it transaction} is a couple $(tid,I)$ where $tid$ is the {\it transaction identifier} and 
$I$ is an itemset. A {\it transaction database} ${\cal D}$ is a finite set of transactions over $\cal I$
where for all two different transactions, they do not have the same transaction identifier.
We say that a transaction $(tid, I)$ {\it supports} an itemset $J$ if  $J\subseteq I$.

The {\it cover} of an itemset $I$ in a transaction database $\cal D$ is the set of
identifiers of  transactions in   $\cal D$ supporting $I$: 
${\cal C}(I,{\cal D})=\{tid\mid (tid,J)\in{\cal D}\texttt{ and }I\subseteq J \}$.
The {\it support} of an itemset $I$ in $\cal D$ is defined by:
${\cal S}(I,{\cal D})=\mid {\cal C}(I,{\cal D})\mid$.
Moreover, the {\it frequency} of $I$ in $\cal D$ is defined by:  
${\cal F}(I,{\cal D})=\frac {{\cal S}(I,{\cal D})}{\mid{\cal D}\mid}$.

For example, let us consider  the transaction database in Table~\ref{tty}.
Each transaction corresponds to the favorite writers of a library member.
For instance, we have ${\cal S}(\{Hemingway,  Melville\},{\cal D})=|\{002,004\}|=2$ and 
${\cal F}(\{Hemingway, $\\ $ Melville\},{\cal D})=\frac{1}{3}$.

\begin{table}
\label{tab:td}
\centering{
{\small
\begin{tabular}{|c|c|}
\hline
tid & itemset \\
\hline
001 &  $Joyce,  Beckett, Proust$	\\
\hline
002 & $Faulkner, Hemingway,  Melville$\\	
\hline
003 &  $Joyce, Proust$\\
\hline
004 & $Hemingway,  Melville$ \\
\hline
005 & $Flaubert, Zola$ \\
\hline
006 & $Hemingway, Golding$ \\
\hline
\end{tabular}}
}
\caption{An example of transaction database ${\cal D}$ }
\label{tty}
\end{table}

\noindent Let $\cal D$ be a transaction database over $\cal I$ and $\lambda$ a minimal support threshold.
The frequent itemset mining problem consists of computing the following set:
${\cal FIM}({\cal D},\lambda)=\{I\subseteq {\cal I}\mid {\cal S}(I,{\cal D})\geq\lambda\}$.

The problem of computing the number of  frequent itemsets is $\# P$-hard~\cite{Gunopulos2003}. The complexity class $\# P$ 
corresponds to the set of counting problems associated with a decision problems in $NP$.
For example, counting the number of models satisfying a CNF formula is a $\# P$  problem.

 
\subsection{Maximal and Closed Frequent Itemsets}
\label{subsec:maxclo}
Let us now define two condensed  representations of the set of all frequent itemsets: 
maximal and closed frequent itemsets. 

\begin{definition}[Maximal Frequent Itemset]
Let $\cal D$ be a transaction database, $\lambda$ a minimal support threshold
and  $I\in{\cal FIM}({\cal D},\lambda)$. $I$ is called maximal when for all $I'\supset I$, 
$I'\notin {\cal FIM}({\cal D},\lambda)$ ($I'$ is not a frequent itemset). 
\end{definition}

We denote by ${\cal MAX}({\cal D}, \lambda)$ the set of all maximal frequent itemsets
in $\cal D$ with $\lambda$ as a minimal support threshold.
For instance, in the previous example, we have ${\cal MAX}({\cal D}, 2)=\{\{Joyce, Proust\},\{Hemingway,  Melville\}\}$. 
 
\begin{definition}[Closed Frequent Itemset]
Let $\cal D$ be a transaction database, $\lambda$ a minimal support threshold
and  $I\in{\cal FIM}({\cal D},\lambda)$. $I$ is called closed when for all $I'\supset I$, 
${\cal C}(I,{\cal D})\neq {\cal C}(I',{\cal D}) $. 
\end{definition}

We denote by ${\cal CLO}({\cal D}, \lambda)$ the set of all closed frequent itemsets
in $\cal D$ with $\lambda$ as a minimal support threshold.
For instance,  we have ${\cal CLO}({\cal D}, 2)=\{\{Hemingway\},$\\$\{Joyce, Proust\},\{Hemingway,  Melville\}\}$.
In particular, let us note that we have ${\cal C}(\{Hemingway\},{\cal D})=\{002,004,006\}$ and ${\cal C}(\{Hemingway,Melville\},{\cal D})=\{002,004\}$. 
That explains why $\{Hemingway\}$ and $\{Hemingway,  Melville\}$ are both closed.
One can easily see that if all the closed (resp. maximal) frequent itemsets are computed, 
then all the frequent itemsets can be computed without using the corresponding database. 
Indeed, the  frequent itemsets correspond to all the subsets of the closed (resp. maximal) frequent itemsets. \\

Clearly, the number of maximal (resp. closed) frequent itemsets is  significantly smaller
than the number of frequent itemsets. Nonetheless, this number is not always polynomial in the size of the database~\cite{Yang2004}. 
In particular, the problem of counting the number of maximal frequent itemsets 
is $\#P$-complete (see also~\cite{Yang2004}).

Many algorithm has been proposed for enumerating frequent closed itemsets. One can cite Apriori-like algorithm, originally proposed in \cite{agrawal93} for mining frequent itemsets for association rules. It proceeds by a level-wise search of the elements of  ${\cal FIM}({\cal D},\lambda)$.
Indeed, it starts by computing the elements of  ${\cal FIM}({\cal D},\lambda)$ of size one.
Then, assuming the element of ${\cal FIM}({\cal D},\lambda)$ of size $n$ is known,  
it computes a set of candidates  of size $n+1$ so that $I$ is a candidate if and only if 
all its subsets are in  ${\cal FIM}({\cal D},\lambda)$. This procedure is iterated until no more candidates are found. 
Obviously, this basic procedure is enhanced using some properties such as the anti-monotonicity property that allow us to reduce the search space. Indeed, if $I\notin{\cal FIM}({\cal D},\lambda)$, then $I'\notin{\cal FIM}({\cal D},\lambda)$ for all  $I'\supseteq I$. In our experiments, we consider one of the state-of-the-art algorithm LCM for mining frequent closed itemsets proposed by Takeaki Uno et al. in \cite{UnoKA04}. In theory, the authors prove that LCM exactly enumerates the set of frequent closed itemsets within polynomial time per closed itemset in the total input size.  Let us mention that LCM algorithm obtained the best implementation award of FIMI'2004 (Frequent Itemset Mining Implementations).

\vspace{-0.4cm}
\section{From CNF Formula to Transaction Database}
\label{sec:cnf2db}
We first introduce the satisfiability problem  and some necessary notations. 
We consider the  conjunctive normal form (CNF) representation for the propositional formulas. A {\em CNF formula}  $\Phi$  is a
conjunction of {clauses}, where a {\em clause} is a disjunction of {literals}. A {\em literal} is a positive ($p$) or negated ($\neg{p}$) 
propositional variable.  The two literals $p$ and $\neg{p}$ are called {\it complementary}. A CNF formula can also be seen as a set of clauses, and a clause as a set of literals. The size of the CNF formula $\Phi$ is defined as $|\Phi| = \sum_{c\in\Phi} |c|$, where $|c|$ is equal to the number of literals in $c$. 
We denote by $\bar{l}$  the complementary literal of $l$. More precisely, if  $l = p$ then $\bar{l}$ is $\neg p$ and if  $l = \neg p$ then $\bar{l}$ is $p$.
Let us recall that any propositional formula can be translated to CNF using Tseitin's linear encoding \cite{Tseitin68}. We denote by ${\cal V}_{\Phi}$ the set of propositional variables appearing in $\Phi$, while the set of literals of $\Phi$ is defined as ${\cal L}_{\Phi}= \cup_{x\in {\cal V}_{\Phi} }\{x,\neg x\}$. An {\it interpretation} ${\cal B}$ of a propositional formula $\Phi$ is a function which  associates a value ${\cal B}(p)\in\{0, 1\}$ ($0$ corresponds to $false$ and $1$ to $true$)
to the variables $p \in {\cal V}_{\Phi}$.  A {\it model} of a formula $\Phi$ is an  interpretation ${\cal B}$ that  satisfies the  formula:  ${\cal B}(\Phi)=1$. 
The {\it SAT problem} consists in deciding if a given CNF formula admits a model or not. 

A CNF formula can be considered as a transaction database, called CNF database,
where the items correspond to literals and the transactions to clauses.  Complementary literals correspond to two different items. 
\begin{definition}[CNF to ${\cal D}$]
Let  $\Phi = \bigwedge_{1\leq i\leq n} c_i$ be a CNF formula. The set of items ${\cal I} = {\cal L}_{\Phi}$ and the transaction database associated to $\Phi$ is defined as ${\cal D}^c_{\Phi} = \{(tid_i, c_i)| 1\leq i\leq n\}$ 
\end{definition}

In this context, a frequent itemset corresponds to a frequent set of literals: the number of clauses containing these literals is greater 
or equal to the minimal threshold. For instance, if we set the minimal threshold $\lambda$ to 2, we get 
$\{x_1, \neg x_2\}$ as a frequent itemset in the previous database. The set of maximal frequent itemsets 
is the smallest set of frequent set of literals where each frequent set of literals is included 
in at least one of its elements. For instance, the unique maximal frequent itemset
in the previous example is  $\{x_1,\neg x_2\}$ ($\lambda=2$).
Furthermore, the set of closed frequent itemsets  is
the smallest set  of frequent set of literals where each frequent itemset is included 
in at least one of its elements {\it having the same support}. For instance, the set of 
the closed frequent itemsets is $\{\{x_1,\neg x_2\},\{x_1\}\}$.

In the definition of a transaction  database, we did not require that the set of items in a transaction 
to be unique. Indeed, two different transactions can have 
the same set of items and different identifiers. A CNF formula may contain the same 
clause more than once, but in practice this does not provide any information about satisfiability.
Thus, we can consider a CNF database as just a set of itemsets (sets of literals).
\vspace{-0.4cm}
\section{Mining-based Approach for Size-Reduction of CNF Formulae}
\label{sec:cr}
In this section, we describe our mining based approach, called Mining4SAT, for reducing the size of CNF formulae.  
The key idea consists in searching for frequent sets of literals (sub-clauses) and substituting them with new variables using Tseitin's encoding~\cite{Tseitin68}.

\subsection{Tseitin's Encoding}

Tseitin's  encoding consists in introducing fresh variables to represent sub-formulae in order  to 
represent their truth values. Let us consider the following DNF formula (Disjunctive Normal Form: a disjunction 
of conjunctions): $$(x_1\lconj\cdots{}\lconj x_l)\ldisj (y_1\lconj\cdots{}\lconj y_m)\ldisj (z_1\lconj\cdots{}\lconj z_n)$$
A naive way of converting such a formula to a CNF formula consists in using the distributivity 
of disjunction over conjunction ($A\ldisj (B\lconj C)\leftrightarrow (A\ldisj B)\lconj (A\ldisj C)$):
$$(x_1\ldisj y_1\ldisj z_1)\lconj (x_1\ldisj  y_1 \ldisj z_2)\lconj\cdots{}\lconj (x_l\ldisj y_m\ldisj z_n) $$
Such a naive approach is clearly exponential in the worst case. In Tseitin's transformation, fresh propositional 
variables are introduced to prevent such combinatorial explosion, mainly caused by the distributivity of disjunction over conjunction and vice versa. With additional variables, the obtained CNF formula is linear in the size of the original formula. However the equivalence is only preserved w.r.t satisfiability:
$$(t_1\ldisj t_2\ldisj t_3)\lconj (t_1\rightarrow (x_1\lconj\cdots{}\lconj x_l)) \lconj (t_2\rightarrow (y_1\lconj\cdots{}\lconj y_m))$$
$\lconj (t_3\rightarrow (z_1\lconj\cdots{}\lconj z_n))$

\subsection{A Size-Reduction Method}

Let us consider the following CNF formula $\Phi$:
$$(x_1\ldisj\cdots\ldisj x_n\ldisj \alpha_1)\lconj\cdots{}\lconj(x_1\ldisj\cdots\ldisj x_n\ldisj \alpha_k)$$
where $n\geq2$, $k> \frac{n+1}{n-1}$, $x_1,\ldots{},x_n$ are literals and $\alpha_1,\ldots{},\alpha_k$ are clauses.
The number of literals in this formula can be reduced as follows:
$$(y\ldisj \alpha_1)\lconj\cdots{}\lconj(y\ldisj \alpha_k)\lconj(x_1\ldisj\cdots\ldisj x_n\ldisj \neg y)$$
where $y$ is a fresh propositional variable. Indeed, $n\times k$ literals are replaced with $k+n+1$ literals. 
Clearly, a boolean interpretation is a model of the formula obtained after reduction if and only if  it  is a model of  $\Phi$.
Now, if we consider the CNF database corresponding to  $\Phi$, $\{x_1,\ldots{},x_n\}$ is a frequent itemset 
where the minimal support threshold is greater or equal to $k$. 

It is easy to see that to reduce the number of literals  $n$ must be  greater or equal to $2$.
Indeed, if $n< 2$ then there is no reduction of the number of literals, on the contrary, their number
is increased.
Regarding the value of $k$, one can also see that such a transformation is interesting only when $k>\frac{n+1}{n-1}$. 
Thus, there are three cases : if $n=2$,  then $k\geq 4$, else if $n=3$ then $k\geq 3$, $k\geq 2$ otherwise. Therefore, 
the number of literals is always reduced when $k\geq 4$.

In the previous example, we illustrate how the problem of finding frequent itemsets can be used to reduce 
the size of a CNF formula. One can see that, in general, it is more interesting to consider a condensed representation 
 of the frequent itemsets (closed and maximal) to reduce the number of literals. Indeed,  
by using a condensed representation, we consider all the frequent itemsets and the number of  fresh propositional variables 
and new clauses (in our example, $y$ and $(x_1\ldisj\cdots\ldisj x_n\ldisj \neg y)$)
introduced  is smaller than  
that of those introduced by using all the frequent itemsets.  For instance, in the previous formula, it is not interesting to introduce 
a fresh propositional variable for each subset of $\{x_1,\ldots{},x_n\}$. 
\vspace{-0.4cm}
\paragraph{\bf Closed vs. Maximal} In Section~\ref{subsec:maxclo}, we introduced two condensed representations of the frequent itemsets: closed and maximal.
The question is, which condensed representation is better?
We know that the set of maximal frequent itemsets is included in that of the closed ones. 
Thus, a small number of fresh variables and new clauses are introduced using the maximal frequent itemsets.
However, there are cases where the use of the closed frequent itemsets is more suitable.
For example, let us consider the following formula:
\begin{center}
\begin{tabular}{c}
$(x_1\ldisj\ldots{}\ldisj  x_k\ldisj  \ldots{}\ldisj  x_n\ldisj \alpha_1)\lconj $\\
$\cdots{}
\lconj(x_1\ldisj\ldots{}\ldisj  x_k\ldisj \ldots{}\ldisj  x_n\ldisj \alpha_m)\lconj$\\
$(x_1\ldisj\ldots{}\ldisj x_k\ldisj\beta_1)\lconj
\cdots{}
\lconj(x_1\ldisj\ldots{} \ldisj x_k \ldisj\beta_{m'})$\\
\end{tabular}
\end{center}
where $k \geq 2$,  $m,m' \geq 4$ and $n > k$.
We assume that the frequent itemsets are only the subsets of $\{x_1,\ldots{},x_n\}$.
Therefore, $\{x_1,\ldots{},x_n\}$ is the unique maximal itemset and 
the  closed itemsets are $\{x_1,\ldots{},x_n\}$ and  $\{x_1,\ldots{},x_k\}$.
Let us  start by using the closed frequent 
itemset $\{x_1,\ldots{},x_n\}$ in the reduction of the number of literals: 
\begin{center}
\begin{tabular}{c}
$(y\ldisj \alpha_1)\lconj 
\cdots{}
\lconj(y\ldisj \alpha_m)\lconj$\\
$(x_1\ldisj\ldots{}\ldisj x_k\ldisj\beta_1)\lconj
\cdots{}
\lconj(x_1\ldisj\ldots{} \ldisj x_k \ldisj\beta_{m'})\lconj$\\
$ (x_1\ldisj \ldots{}\ldisj x_n\ldisj \neg y)$
\end{tabular}
\end{center}
Now, by using  $\{x_1,\ldots{},x_k\}$, we get the following formula:
\begin{center}
\begin{tabular}{c}
$(y\ldisj \alpha_1)\lconj 
\cdots{}
\lconj(y\ldisj \alpha_m)\lconj$\\
$(z\ldisj\beta_1)\lconj
\cdots{}
\lconj(z \ldisj\beta_{m'})\lconj$\\
$ (z  \ldisj x_{k+1}\ldisj \ldots{}\ldisj x_n\ldisj \neg y) \lconj (x_1\ldisj\ldots{} \ldisj x_k \ldisj \neg z)$
\end{tabular}
\end{center}
In this example, it is clearly more interesting to consider  the closed frequent itemsets in our Mining4SAT approach.

In fact, a (closed) frequent itemset $I$ and one of its subsets $I'$ (which can be closed) are both interesting 
if ${\cal S}(I')-{\cal S}(I)> \frac{|I'|+1}{|I'|-1}-1$. Indeed, if we apply our transformation using $I$, 
then the support of $I'$ in the resulting formula is equal to ${\cal S}(I')-{\cal S}(I)+1$, and we know 
that $I'$ is interesting in the resulting formula if its support is greater to  $ \frac{|I'|+1}{|I'|-1}$.

\paragraph{\bf Overlap} Let  $\Phi$ be a set of  itemsets.
Two  itemsets $I$ and $I'$ of  $\Phi$ overlap if $I\cap I'\neq \emptyset$.
Moreover, $I$ and $I'$ are in the same {\it overlap class} if there exist $k$ itemsets $I_1,\ldots{},I_k$ of  $\Phi$  
such that  $I=I_1, I_k=I'$  and for all $1\leq i\leq k-1$,  $I_i$ and $I_{i+1}$ overlap.

In our transformation, one can have some problems when two frequent itemsets overlap. 
For example, if $\{x_1,x_2,x_3\}$ and $\{x_2,x_3,x_4\}$ are two frequent itemsets (3 is the minimal support threshold) such that 
${\cal S}(\{x_1,x_2,x_3\})=3$, ${\cal S}(\{x_2,x_3,$\\$x_4\})=3$ and   ${\cal S}(\{x_1,x_2,x_3,x_4\})=2$, 
then if we apply our transformation using   $\{x_1,x_2,x_3\}$, then  the support of $\{x_2,x_3,x_4\}$  is 
equal to $2$ (infrequent) in the resulting formula  
and vice versa. Thus, we can not use both of them in the transformation.

Le us note that the overlap notion can be seen as  a generalization of the subset one.
Let $I$ and $I'$ be frequent itemsets such that they overlap. 
They are both interesting in our transformation if:
\begin{enumerate}
\item   ${\cal S}(I)-{\cal S}(I\cup I')> \frac{|I|+1}{|I|-1}-1$
or ${\cal S}(I')-{\cal S}(I\cup I')> \frac{|I'|+1}{|I'|-1}-1$.
This comes from the fact that if we apply the transformation using $I$ (resp. $I'$), 
then the support of $I'$ (resp. $I$) is equal to ${\cal S}(I')-{\cal S}(I\cup I')+1$
(resp. ${\cal S}(I)-{\cal S}(I\cup I')+1$). \\

\item  $|I\backslash I'|\geq k$ (resp. $|I'\backslash I|\geq k$) where 
$k=2$ if ${\cal S}(I)\geq 4$ (resp. ${\cal S}(I')\geq 4$), 
$k=3$ if ${\cal S}(I) = 3$ (resp. ${\cal S}(I')= 3$),
$k=4$ otherwise. Indeed, in the previous cases, $I\backslash I'$ (resp. $I'\backslash I$)
can be used in our transformation.
\end{enumerate}
\vspace{-0.4cm}
\paragraph{\bf Mining4SAT algorithm}

We now describe our Mining4SAT algorithm using the set of closed frequent itemsets. 
Let us note that the optimal transformation using the set of all the closed frequent itemsets can be obtained 
by an optimal transformation using separately the overlap classes of this set. Actually, 
since  any two distinct overlap classes do not share any literal, the reduction applied  
to a given formula using the elements of an overlap class does not affect the supports of the elements 
of the other classes. Moreover, one can easily compute the set of all the overlap classes of
the set of the closed frequent itemsets: let $G=(V,E)$ be an undirected graph such that 
$V$ is the set of the closed frequent itemsets and $(I_1,I_2)$ is an edge of $G$ if and only if 
$I_1$ and $I_2$ overlap; $C$ is an overlap class if and only if it corresponds to the set of  vertices 
of a connected component of $G$ which is not included in any other connected component of $G$.
For this reason, we restrict here our attention to the reductions that can be obtained using a single overlap class. The hole size reduction process can be performed by iterating on all the overlap classes.

Let  $I$ be a closed frequent itemset,
We denote by $\alpha(I)$ the value ${\cal S}(I)\times (|I|-1)-|I|-1$ that corresponds 
to the number of literals reduced by applying our transformation with $I$ on a CNF formula. 

Algorithm~\ref{algo1} takes as input a CNF formula $\phi$ and an overlap class $C$,
and returns $\phi$ after applying size-reduction transformations.
It iterates until there is no element in $C$. In each iteration, it first selects one of the 
most interesting elements  in $C$ (line 2): an element $I$ of $C$ such that there is no element  $I'\in C$  
satisfying $\alpha(I') > \alpha(I)$. Note that this element is not necessarily unique in $C$.
This instruction means that Algorithm~\ref{algo1} is a greedy algorithm because it makes the locally optimal choice 
at each iteration.  
Then, it applies our transformation using $I=\{y_1,\ldots{},y_n\}$: it replaces the occurrences of $I$ with 
a fresh propositional variable $x$ (line 3); and it adds the clause $y_1\vee\ldots{}\vee y_n\vee\neg x$
to $\phi$ (line 4). It next removes $I$ from $C$ (line 5) and replaces $I$ in the the other elements of $C$ with $x$ (line 6).
The next instruction (line 7) consists in removing the elements of $C$
that could increase the number of literals: the elements that overlap with $I$ 
and are not included in $I$.
As explained before, an element of $C$ overlapping with $I$ does not necessarily increase 
the number of literals.  Thus, by removing  elements from $C$ because only they overlap with $I$,
our algorithm can remove closed frequent itemsets decreasing the number of literals. 
A partial solution to this problem consists in recomputing the closed frequent itemsets 
in the formula returned by Algorithm~\ref{algo1}. 
The last instruction  in the while loop (line 8) consists in updating the supports 
of the elements remaining in $C$ following the new value of $\phi$: 
a support of an element $I'$ remaining in $C$ changes only when it is included in $I$
and its new support is equal to ${\cal S}(I')-{\cal S}(I)+1$. This instruction also removes 
all the elements of $C$ becoming uninteresting because of the new supports and sizes.
\begin{algorithm}[t]
\caption{Size Reduction}
\label{algo1}
\label{algo_rrt}
\begin{algorithmic}[1]
\REQUIRE A formula $\phi$, an overlap class of closed frequent itemsets $C$
\WHILE {${C}\neq \emptyset$}
   \STATE  $I \leftarrow MostInterstingElment(C)$;
   \STATE  $replace(\phi, I,x)$;
   \STATE $Add(\phi, I,x)$:
   \STATE $ remove(C, I)$;
   \STATE $ replaceSubset(C, I,x)$;
   \STATE $ removeUninterestingElements(C)$;
   \STATE $ updateSupports(C)$;
\ENDWHILE
\RETURN $\phi$
\end{algorithmic}
\end{algorithm}
\vspace{-0.45cm}
\section{Application: A Compact Representation of Sets of Binary Clauses}
\label{sec:bin}

Binary clauses (2-CNF formula) are ubiquitous in CNF formula encoding real-world problems. Some of them contain more than 50\% of binary clauses. However, in our size reduction approach, binary clauses are not taken into account.  Indeed, to reduce the size of the formula, we only search for itemsets of size at least two literals.  The  extremely rare case where a binary clause representing a closed frequent itemset can be considered is when it appears at least four times in the formula i.e. it subsumes at least 4 clauses.  In this section, we first show how our mining based approach can be used to achieve a compact representation of arbitrary sets of binary clauses.  Then, we consider two interesting special cases corresponding to sets of binary clauses representing either a clique or a bi-clique. 
\subsection{Compacting arbitrary set of binary clauses} 
In order to reduce the size of the set of binary clauses, we only need to rewrite the formula and to slightly modify the Algorithm \ref{algo1}. 

\begin{definition}[B-implications]
Let  $\Phi = \bigwedge_{1\leq i\leq n} [(x_i\vee y^i_1)\wedge(x_i\vee y^i_2)\wedge\dots\wedge(x_i\vee y^i_{n_i})]$ be a 2-CNF formula. We define $B_{\vee[\wedge]}(\Phi) = \bigwedge_{1\leq i\leq n}x_i\vee\beta_i$, where $\beta_i=(y^i_1\wedge y^i_2\wedge\dots\wedge y^i_{n_i})$. We call $(x_i\vee\beta_i)$ a  {\it B-implication}. 
\end{definition}

Obviously, the formula $\Phi$ and $B_{\vee[\wedge]}(\Phi)$ are equivalent and there exists several ways to rewrite $\Phi$ as a conjunction of B-implications. 
\begin{example}
\label{ex:bi}
Let $\Phi = (a\vee b)\wedge (a\vee c)\wedge (c\vee d)$ be a 2-CNF formula. We can rewrite $\Phi$ as  $B^1_{\vee[\wedge]}(\Phi) = (a\vee [b\wedge c])\wedge (c\vee [d])$ or as $B^2_{\vee[\wedge]}(\Phi) = (a\vee [b]) \wedge (c\vee[a\wedge d])$. 
\end{example}
In the sequel, we use a lexicographic ordering on literals of $\Phi$. In the example \ref{ex:bi}, we obtain  $B^1_{\vee[\wedge]}(\Phi)$ using the lexicographic ordering  $\neg a\leq a <\neg b\leq b < \neg c\leq  c < \neg d\leq d$.

\begin{definition}[2-CNF to ${\cal D}$]
Let  $\Phi$ be a 2-CNF formula and $B_{\vee[\wedge]}(\Phi) = \bigwedge_{1\leq i\leq n}x_i\vee\beta_i$. The transaction database associated to $\Phi$ is defined as ${\cal D}^b_{\Phi} = \{(tid_i, \beta_i)| x_i\vee\beta_i \in B_{\vee[\wedge]}(\Phi)\}$. 
\end{definition}

Let us now describe our approach to compact a 2-CNF formula $\Phi$, called CNF2RED (for reducing the size of sets of binary clauses).  First, after rewriting $\Phi$ as  $B_{\vee[\wedge]}(\Phi)$, we build the transaction database ${\cal D}^b_{\Phi}$.  The set ${\cal C}iSet$ of closed frequent itemsets and its associated overlap classes ${\cal O}class$ are computed. The last step aims to reduce the size of the 2-CNF $\Phi$ using a slightly modified version of the Algorithm \ref{algo1}. First the Algorithm \ref{algo1} takes as input a formula $\phi = B_{\vee[\wedge]}(\Phi)$ and returns $\phi$ after reducing its size. Secondly, for an itemset $I = \{y_1,y_2,\dots,y_n\}$, in line (4) of the Algorithm \ref{algo1}, we introduce a fresh variable $x$ and we add a bi-implication $(\neg x\vee [y_1\wedge y_2\wedge\dots\wedge y_n])$ to $\phi$. 

\subsection{Special case of (bi-)clique of binary clauses}
In  \cite{Rintanen06},  J. Rintanen addressed the problem of representing big sets of binary clauses compactly. He particularly shows that constraint graphs arising from practically interesting applications  (eg. AI planning) contain big cliques or bi-cliques of binary clauses.  An identified bi-clique involving the two sets of literals $\mathcal{X} = \{x_1, x_2, \dots, x_n\}$ and $\mathcal{Y} = \{y_1, y_2, \dots, y_m\}$ expresses the propositional formula $(x_1\wedge x_2\wedge\dots\wedge x_n) \vee (y_1\wedge y_2\wedge\dots\wedge y_m)$, while a clique involving the literals $\mathcal{X} = \{x_1, x_2, \dots, x_n\}$ expresses that at-most one literal from $\mathcal{X}$ is $false$, 
\vspace{-0.4cm}
\paragraph{Bi-clique of binary clauses}
Let us explain how a bi-clique can be compacted with CNF2RED method. Let $\Psi = [(x_1\vee y_1)\wedge (x_1\vee y_2)\vee\dots\vee (x_1\vee y_m)]\dots [(x_n\vee y_1)\wedge (x_n\vee y_2)\vee\dots\vee (x_n\vee y_m)] $ a bi-clique of $n\times m$ binary clauses.  Considering the lexicographic ordering, $B_{\vee[\wedge]}(\Psi)$ corresponds exactly to $\bigwedge_{1\leq i\leq n} (x_i\vee[y_1\wedge y_2\wedge\dots\wedge y_m])$. Obviously,  the transaction database ${\cal D}^b_{\Psi}$ contains a single closed frequent itemset $\{y_1,y_2,\dots,y_m\}$. Applying our algorithm leads to the following compact representation of $\Psi' = [\bigwedge_{1\leq i\leq n} (x_i\vee z)]\wedge[\bigwedge_{1\leq j\leq m} (\neg z\vee y_j)]$. We obtain exactly the same gain as in \cite{Rintanen06} (${\cal O}(n+m)$ binary clauses and one additional variable). 
\vspace{-0.4cm}
\paragraph{Clique of binary clauses} Let $\Psi = \bigwedge_{1\leq i\leq n-1} [(x_i\vee x_{i+1})\wedge (x_i\vee x_{i+2})\vee\dots\vee (x_i\vee x_n)]$ 
be a clique of $n^2$  binary clauses. The formula $B_{\vee[\wedge]}(\Psi) =  \bigwedge_{1\leq i\leq n-1} (x_i\vee [ x_{i+1}\wedge x_{i+2}\wedge\dots\wedge x_n])$. 
If we take a closer look to  $D^b_{\Psi}$, the closed frequent itemset $I$ with greatest value $\alpha(I)$ corresponds to $\{x_{n/2},\dots,x_n\}$. 
In the first $\frac{n}{2}$ rows of $D^b_{\Psi}$, $I$ is substituted by a fresh variable $x$ and a new  set of binary clauses $(x\vee [x_{\frac{n}{2}}\wedge, \dots\wedge x_n])$ is added to it, leading to two subproblems 
of size $\frac{n}{2}+1$. Obviously, the same treatment is done on the formula $B_{\vee[\wedge]}(\Psi)$. Consequently the number of variables is defined by the following recurrence equation: ${\cal V}(n) = 2 {\cal V}(\frac{n}{2}+1) +1$,  ${\cal V}(6) =1$. The basic case is reached for $n=6$, where the last fresh variable is introduced to represent the conjunction $x_4\wedge x_5\wedge x_6$. For $n< 6$ no fresh variable is introduced because no frequent closed itemset can leads to a reduction of the size of the formula. Consequently, from the solution of the previous recurrence equation, we obtain that our encoding is in ${\cal O}(n)$ auxiliary variables. Using the same reasoning, we also obtain the same complexity ${\cal O}(n)$ for the number of binary clauses. This corresponds to the complexity obtained in \cite{Rintanen06}.   

The two special cases of clique and bi-clique of binary clauses considered in this section, allow us to show that when a constraint is not well encoded, our approach can be used to correct and to derive a more efficient and compact encodings automatically. 

\section{Experiments}
\label{sec:expes}
\begin{table}[htbp]
{\scriptsize
\centering
\begin{tabular}{|l|l|l|l|}
\hline
{\bf Instance} &  {\bf orig. form. size}  & {\bf red. form. size}   & {\bf \% rmv}\\
\hline 

 1dlx\_c\_iq57\_a & 190 Mo  & 164 Mo  &  12,47 \% \\
 6pipe\_6\_ooo.*-as.sat03-413 & 11 Mo  & 7,7 Mo  &  19,64 \% \\
 9dlx\_vliw\_at\_b\_iq6.*-*04-347 & 76 Mo  & 65 Mo  &  14,02 \% \\
 abb313GPIA-9-c.*.sat04-317 & 21 Mo  & 6,9 Mo  &  63,92 \% \\
 E05F18 & 3,7 Mo  & 2,2 Mo  &  43,48 \% \\
eq.atree.braun.11.unsat & 120 Ko  & 72 Ko  &  27,93 \% \\
eq.atree.braun.12.unsat & 144 Ko  & 88 Ko  &  27,66 \% \\
 k2mul.miter.*-as.sat03-355 & 1,5 Mo  & 1,3 Mo  &  11,27 \% \\
korf-15 & 1,2 Mo  & 752 Ko  &  34,17 \% \\
rbcl\_xits\_08\_UNSAT & 1,1 Mo  & 856 Ko  &  16,42 \% \\
 SAT\_dat.k45 & 3,5 Mo  & 2,6 Mo  &  24,53 \% \\
 traffic\_b\_unsat & 18 Mo  & 12 Mo &  26,53 \% \\
x1mul.miter.*-as.sat03-359 & 1,1 Mo  & 928 Ko  &  12,68 \% \\
 9dlx\_vliw\_at\_b\_iq3  & 19 Mo  & 15 Mo  &  17,84 \% \\
9dlx\_vliw\_at\_b\_iq4  & 31 Mo  & 26 Mo  &  18,02 \% \\
 AProVE07-09  & 2,8 Mo  & 2,7 Mo  &  4,51 \% \\
 eq.atree.braun.10.unsat  & 96 Ko  & 56 Ko &  28,30 \% \\
 goldb-heqc-frg1mul  & 348 Ko & 328 Ko  &  12,66 \% \\
 goldb-heqc-x1mul  & 964 Ko  & 896 Ko  &  12,68 \% \\
minand128  & 7,7 Mo  & 2,6 Mo &  65,28 \% \\
ndhf\_xits\_09\_UNSAT  & 2,6 Mo  & 2,1 Mo  &  18,61 \% \\
rbcl\_xits\_07\_UNSAT  & 868 Ko  & 720 Ko  &  16,49 \% \\
 velev-pipe-o-uns-1.1-6  & 5,5 Mo  & 4,4 Mo  & 18,89 \% \\

\hline
\end{tabular}
}
\caption{Results of Mining4SAT : a general approach}
\label{tab:cnf}
\end{table}


In this section, we present an experimental evaluation of our proposed approaches. Two kind of experiments has been conducted. The first one deals with size reduction of arbitrary CNF formulas using Mining4SAT algorithm, while the second one attempts to reduce the size of the 2-CNF sub-formulas only, using CNF2RED algorithm.

Both algorithms are tested on different benchmarks taken from the last SAT challenge 2012. From the 600 instances of the application category submitted to this challenge, we selected 100 instances while taking at least one instance from each family. All tests were made on a Xeon 3.2GHz (2 GB RAM) cluster and the timeout was set to 4 hours. 
 
In Table \ref{tab:cnf} and Table \ref{tab:bin}, the field $size$ indicates the size in
octets of each SAT instance before and after reduction. We also provide $\% rmv$, the percentage of the removed literals. To study  the influence of  our size reduction approaches on the solving time, we also run the SAT solver MiniSAT 2.2 on both the original instance and on the those obtained after reduction.  Due to a lack of space, we only present a sample of the whole results. Our goal is to provide some insights about the general behavior of our reduction techniques. 

Table \ref{tab:cnf}, highlights the results obtained by Mining4SAT general approach. In this experiments, and to allow possible reductions, we only search for frequent closed itemsets of size greater or equal to 4. Consequently, binary clauses are not considered. As we can observe, our Mining4SAT reduction approach allows us to reduce the size more than 20\% on the majority of instances. Let us also note that  the maximum (65,28 \%) is reached in the case of  the instance {\it minand128}: 
its original size is  {\tt 14 Mo} and its size after reduction is {\tt 5.4 Mo}. For the SAT solving time, the results depend on the instances. On some instances we can observe real improvements, whereas on others the performances become worse. 

In Table \ref{tab:bin}, we present a sample of the results obtained by CNF2RED algorithm on compacting only binary clauses. We observe similar behavior as in the first experiment in terms of size reduction However, we observe in general some improvements in terms of SAT solving time.

\begin{table}[htbp]
{\scriptsize
\centering
\begin{tabular}{|l|l|l|l|}
\hline
{\bf Instance} &  {\bf orig. form. size}  & {\bf red. form. size}   & {\bf \% rmv}\\
\hline 
velev-pipe-o-uns-1.1-6  & 5.5 Mo     & 3.2 Mo   & 43,23 \% \\
\hline
9dlx\_vliw\_at\_b\_iq2       &    11 Mo            & 6 Mo  &  42,56 \% \\
\hline
1dlx\_c\_iq57\_a                  &     190 Mo        &  124 Mo  & 36,52 \% \\
\hline
 7pipe\_k                               &  14 Mo            & 5.4 Mo    & 59,66 \% \\ 
\hline
SAT\_dat.k100.debugged  & 16 Mo               &  13 Mo      & 24,89 \% \\
\hline
IBM\_FV\_2004\_rule\_batch &  9,7 Mo    &  7.5 Mo   &  25,56  \% \\
\_2\_31\_1\_SAT\_dat.k80.debugged   & & & \\
\hline
sokoban-sequential-p145-*.040-* & 24 Mo  &  14 Mo   & 45,16  \%\\
\hline
openstacks-*-p30\_1.085-* &30 Mo   & 26 Mo    &  17,25 \%   \\
\hline
aaai10-planning-ipc5-*-12-step16 & 17 Mo     &  12 Mo  &  35,35 \%  \\
\hline
k2fix\_gr\_rcs\_w8.shuffled & 3,4 Mo   & 1,7 Mo   &   54,83\%\\
\hline
homer17.shuffled & 20 Ko   & 16 Ko   &   39,86 \% \\
\hline
gripper13u.shuffled-as.sat03-395  &524 Ko  &   364 Ko  & 35,03 \% \\
\hline
grid-strips-grid-y-3.045-*  &  52 Mo  & 42 Mo     & 23,48 \%\\
\hline
\end{tabular}
}
\caption{Results of CNF2RED: a 2-CNF approach}
\label{tab:bin}
\end{table}
  
%

\section{Conclusion and Future Works}
In this paper, we propose the first data-mining approach, called Mining4SAT, for 
reducing the size of Boolean formulae in conjunctive normal form (CNF).
It can be seen as a preprocessing step that aims to discover hidden structural knowledge
that are used to decrease the number of literals.
Mining4SAT combines both frequent itemset mining techniques for discovering interesting 
substructures, and Tseitin-based approach for a compact representation of CNF formulae 
using these substructures. Thus, we show in this work,  inter alia, that frequent itemset
mining techniques are  very suitable for discovering interesting patterns in CNF formulae.

Since we use a greedy algorithm in our approach, the formula obtained after transformation 
is not guaranteed to be optimal w.r.t. size. An important open question, which we will study in future work, 
is how to optimally use the closed frequent itemsets ranging in an overlap class. Integrating the reduction of 
sets of binary clauses in the general Mining4SAT approach is also an interesting research perspective. 


\bibliographystyle{plain}
\bibliography{biblio}
\end{document}